# A Blind navigation method in indoors/outdoors areas

Mohammad Javadian Farzaneh [a*], Hossein Mahvash Mohammadi [a*]

[a] *Department of Computer Engineering, University of Isfahan, Isfahan, Iran*
[*] m.javadian@eng.ui.ac.ir , h.mahvash@eng.ui.ac.ir

## Abstract:

According to WHO statistics, the number of visually impaired people is increasing annually. One of the most critical necessities for visually impaired people is the ability to navigate safely. This paper proposes a navigation system based on the visual slam and Yolo algorithm using monocular cameras. The proposed system consists of three steps: obstacle distance estimation, path deviation detection, and next-step prediction. Using the ORB-SLAM algorithm, the proposed method creates a map from a predefined route and guides the users to stay on the route while notifying them if they deviate from it. Additionally, the system utilizes the YOLO algorithm to detect obstacles along the route and alert the user. The experimental results, obtained by using a laptop camera, show that the proposed system can run in 30 frame per second while guiding the user within predefined routes of 11 meters in indoors and outdoors. The accuracy of the positioning system is 8cm, and the system notifies the users if they deviate from the predefined route by more than 60 cm.

## Introduction:

According to WHO statistics, approximately 2.2 billion people currently experience visual impairments, a number that is expected to triple by 2050. The blind people require methods to comprehend their environment and navigate safely. Numerous research studies have focused on proposing navigation systems and obstacle detection methods for individuals with visual impairments. Visual impairments do not necessarily hinder navigation, as individuals can utilize their devices to navigate independently and they can rely on their memory and verbal descriptions [1] [2].

An efficient blind navigation system typically consists of three steps, including user localization, goal recognition, and determining the route to the destination. The last step includes way-finding, route following, and obstacle detection [4]. Navigation systems face several challenges including limited preview, insufficient knowledge of the surroundings, and limited access to positioning information. Researchers are trying to develop technological systems to overcome these challenges [5].

Researchers have proposed several solutions for safe and effective blind navigation systems. Additionally, certain institutes provide orientation and mobility training to visually impaired individuals (VIPs) [2],[6]. Orientation refers to understanding one's location and desired destination, while mobility involves traveling without difficulties and dangers [7]. Proposed solutions include white canes, trained volunteers, paved sidewalks and public spaces, and guide dogs. Guide dogs are considered highly effective, but they can be costly and time consuming, and there are limited regulations governing their use [6].

Available assistance and devices for VIPs can be categorized into two groups: The first category includes widely used tools such as white canes or guide dogs, while the second category includes products utilizing Assistive Technology [8]. This technology includes all the devices, services, and processes that help disabled people to overcome physical and social problems [9].

There are some classifications for blind-related assistive technologies, especially in blind navigation [10]. However, these classifications often have overlapping boundaries. Kuriakose et al. [2] have divided blind assistive systems into three categories: Electronic Orientation Aid (EOA), Position Locator Devices (PLD), and Electronic Travel Aids (ETA).

EOAs primarily help users in pathfinding and obstacle detection through the use of cameras and sensors. One disadvantage of these devices is their limited compatibility with other real-time guiding devices, which adds complexity to the system when integrating them.

PLD devices are designed to locate the positions of VIPs using GPS (Global Positioning System) or GIS (Geographic Information System) systems. However, these devices are not functional in indoor areas and require additional sensors to detect obstacles and notify users.

ETAs serve as general assistant devices that aid VIPs in safe navigation and offer a wider range of obstacle detection. They provide better orientation for users compared to other categories. Typically, ETAs incorporate sensors to gather input from the environment and feedback modules to inform users about the next steps.
Chaudary et al. [8] have classified blind assistive systems into two categories: ETAs and Orientation and Navigation Systems (ONS). According to this classification, ETAs gather data from the environment to help users in mobility and obstacle detection, whereas ONS utilize sensors like GPS to help users navigate and reach their intended destinations. Assistive navigation systems [2] were classified into five categories: Visual Imagery Systems, Non-Visual Data Systems, Map-Based Systems, Systems With 3D Sounds, and Smartphone-Based Solutions [2].

GPS-based systems suffer from low accuracy and significant delay in complex and indoor areas due to weak signal strength [11]. Other alternatives include Radio Frequency Identification (RFID), Bluetooth beacons, and infrared sensors [2]. Implementing these systems requires modification and preparation of the environment, which is costly and time-consuming for outdoor areas but can be an option for indoors[2],[12]. Other alternatives are SLAM and V-SLAM techniques which are practical solutions to this problem [11].

The contribution of this paper is related to Visual Imagery Systems (VIS). These systems utilize machine vision algorithms and optical sensors to extract and process environmental features, enabling navigation and obstacle detection through Simultaneously Localization and Mapping (SLAM). VIS incorporates various devices such as monocular/stereo/RGB-D or Lidar devices and frameworks like SLAM, including Visual SLAM (V-SLAM) [2].

This paper is organized as follows. An overview of the blind navigation systems is presented in the next section. Section three presents the proposed method. The last section shows the results and concludes the paper.

## Background Work:

Blind navigation systems can be categorized based on the localization algorithm, whether they use GPS or other localization sensors, if they incorporate a visual positioning system, and if they are functional in indoors and outdoors areas. This section, presents a review of blind navigation systems.

More et al. [13] proposed a smart cane which has nearly ten sensors attached to a Raspberry Pi, a camera to detect objects, and an ultrasonic sensor to measure the distance of obstacles. It utilizes GPS to calculate the possible route and an RF receiver to determine the location of the cane. Additionally, the cane includes water and fire sensors to enhance safety.

Díaz-toro et al. [14] proposed a system to guide a VIP in unfamiliar areas. They introduced a four-step algorithm that incorporates autonomous driving concepts. In the first step, real-time ground segmentation is performed using gravity vectors to find 3D points of obstacles and ground-related areas. Subsequently, the outliers are removed using RANSAC [15]. In the second step, these points are utilized to generate a 2D occupancy grid. This grid is a 2D gray-scale image, where obstacles are depicted in black and free spaces in white. In the third step, the user's position is indicated within the occupancy grid by a semi-circle divided into six equal pieces with a radius of r. In the last step, if obstacles are placed in one or more of these pieces the system notifies the user with specific vibration pattern using a haptic belt. This system helps VIPs avoid obstacles but does not provide navigation information or pathfinding solutions [4].

CHAUDARY et al. [8] proposed a system that utilizes images for pathfinding and navigation. This system comprises a VIP terminal and a remote caretaker terminal. The VIP terminal includes an augmented cane with a vibration-based interface to notify the user. Additionally, the user sends real-time images to the TeleNavigation app on a smartphone, which is accessed from the caretaker terminal .At the caretaker terminal, a person assists VIPs with navigation and way-finding through voice commands.

A key technology that researchers use to develop navigation systems is SLAM, especially V-SLAM. BAI et al. [4] proposed a system using V-SLAM. This system includes Visual SLAM, PoI-graph, way-finding, route following, and obstacle detection. The concepts of route following and obstacle detection are derived from their previous work [16]. The authors used ORB-SLAM 2 [17] to implement V-SLAM for map creation and VIP localization. Images of an area are captured by a sighted individual and then provided to ORB-SLAM for map creation to facilitate VIP localization. Subsequently, points of interest such as rooms, toilets, and hallway junctions are identified using the PoI-graph on the map. Each section of the map represents an edge connecting two points of interest, with the weight of the edges indicating the actual distance between nodes. The A* algorithm is applied to the PoI-graph within the wayfinding module to find the shortest path.

The next step is to process the route-following module. This module gets a set of inputs, including the shortest path, the user's current location, and navigation commands based on obstacles. The output is data that ensures VIP is on the correct and safe route. This system only supports indoor navigation.

Khan et al. proposed a system that supports both indoor/outdoor areas [5]. This system comprises four main parts: a helmet with stereo cameras, a smartphone, a cloud processing platform, and a web application. The helmet is equipped with a stereo camera to capture images, a microphone, and a speaker to receive voice commands. To send data to the cloud, VIP uses a smartphone. The cloud platform consists of three parts: Perception, Navigation, and Speech processing. The Perception component is responsible for object detection tasks. The Navigation component utilizes V-SLAM algorithms to create a map, and the last part employs RNN algorithms to process speech commands.

Another application of V-SLAM utilizes RGB-D cameras to create a 2D map [18]. The proposed framework has four parts of Object detection [19], V-SLAM, and Path planning/following. For the SLAM part, they have used Hector SLAM [20]. To show the map, they used a 2D occupancy grid. The system shows free cells using white and occupied cells using black color. To create an occupancy grid, they used Otsu's thresholding algorithm [21]. For path planning, they used the A* algorithm.

Chen et al. proposed another example of V-SLAM based on ORB features [6]. They concluded that ORB is the best choice for real-time applications compared to other descriptors such as SIFT, SURF, or BRIEF. Moreover, they employed a semantic segmentation method that works in conjunction with V-SLAM. This part is a modified version of ENet [22]. ENet is a small training model consists of 34 residual layers to speed up the training phase. After creating the map using V-SLAM and image segmentation, a data association is used to bring semantic information into the three-dimensional space of the map.

One problem with GPS devices is the time-consuming process of finding the user's orientation and precise location. Duh et al. [23] proposed a method to address this issue using V-SLAM, which can support both indoor and outdoor areas. The proposed system combines V-SLAM with model-based localization methods (MBL) [24] to achieve high accuracy and drift-free positioning. For the V-SLAM component, ORB-SLAM is utilized. Initially, the camera sends images to ORB-SLAM to obtain relative pose estimation. The keyframe is then sent to the MBL and semantic segmentation modules. The outputs of V-SLAM and MBL together define the accurate location of the user. Tracking loss is a common challenge in V-SLAM solutions. To mitigate this issue, the system continuously stores the camera pose on the server to be used in case of camera loss. For semantic segmentation, FRRN [25] is employed, although it does not achieve object-level precision as the system may occasionally miss some obstacles in the area. Instead, semantic segmentation is applied at the pixel level to identify most of the obstacles.

## Proposed Methods:

In this paper, we proposed a system that helps VIPs to navigate easily in the predefined map environment. The goal is to facilitate the movement of VIPs in the environments where they have the most daily traffic. The proposed system consists of three major parts of Map Creation, Path Following, and Obstacle Detection which are explained in detail as follows.

An overview of the proposed system is depicted in Figure 1. Initially, the system checks the availability of the predefined map. If the map is unavailable, the system generates the map by utilizing the ORB-SLAM algorithm and processing the captured video along the route. In case the map is already available, the user can select between two options: Online and Offline tracking.

The ORB-SLAM algorithm consists of three main threads: tracking, local mapping, and loop closing, which operate synchronously. The tracking thread extracts feature points to estimate the relative camera poses for each frame. However, not all feature points are selected, as many of them are duplicates or redundant.

The local mapping thread processes the feature points of selected keyframes and converts them into map points. Keyframes are chosen based on a policy within the tracking thread. Although an initial map estimation is obtained through the execution of the first two threads, it may contain errors, such as accumulated drift, and therefore requires further refinement for improved accuracy.

The loop closing thread is responsible for identifying loops, reducing errors, and creating consistent maps. It also enables ORB-SLAM to reallocate the camera if tracking is lost, thanks to the loop closing process.

Offline tracking is provided to allow users to verify the correctness of the map. The tracking thread is enabled in this mode, while the two other threads are disabled. In online mode, a monocular camera mounted on the glasses of a blind person captures video frames from the scene ahead. These frames are then processed by the obstacle detection and path-following algorithms. The obstacle detection is implemented using YOLOv4. The proposed path following method comprises three steps: obstacle distance estimation, path deviation detection, and next step prediction, as illustrated in Figure 1.

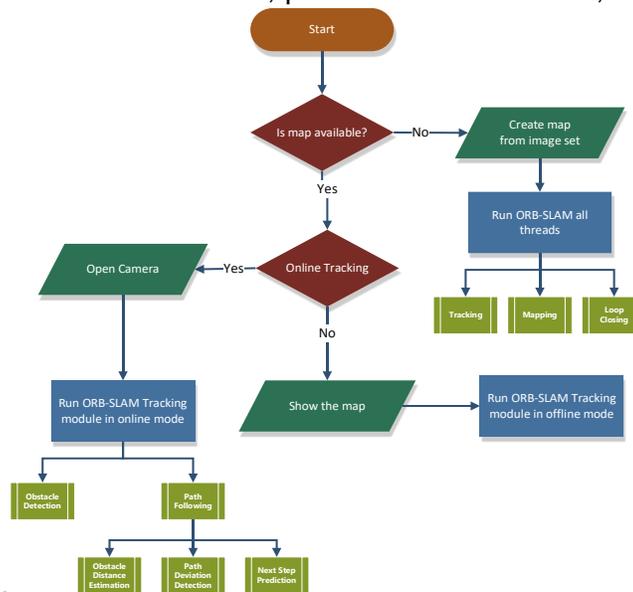

*Figure 1, VSLAM Overview [28]*

## Map Creation:

The SLAM algorithm is used to create a map from the user's route and environment. SLAM is a computer vision technology that creates a map from an environment and estimates the current location. It plays a crucial role in the advancement of self-driving cars, pathfinding, and indoor/outdoor map creation. SLAM can be implemented using various sensors such as Cameras, IMU, GNSS antennas, and Lidars[26]. The proposed method uses a camera as a sensor in SLAM which is called visual SLAM or V-SLAM [26][27].Visual SLAM algorithms primarily consist of two sections: the frontend and backend. Figure 2 illustrates the block diagram of the V-SLAM algorithm.

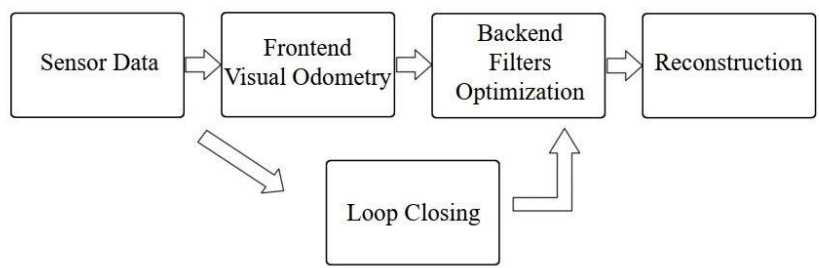

*Figure 2, VSLAM Overview [28]*

The frontend section consists of a Visual Odometry, which estimates camera motion and creates a local map using consecutive images. The backend and loop closing sections are responsible to update the estimated values to achieve map consistency. The main goal of the backend is map or camera trajectory estimation, aiming to measure and improve uncertainty. The loop closing section detects observed trajectories through image matching [28].

The visual odometry estimates the camera position and orientation and the map. However, if the subsequent steps are not executed properly, the estimation may contain errors such as accumulated drift, or the system may become trapped in an infinite map if it does not recognize that the camera has revisited a previous location.

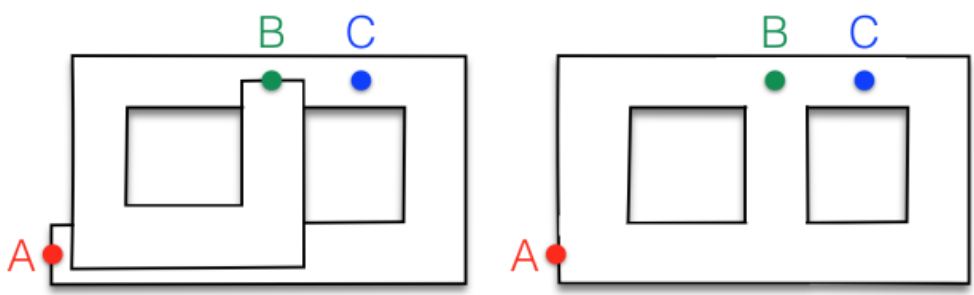

*Figure 3: Left image result of VO, Right image result of SLAM [29]*

As shown in figure 3, If loop closing is not applied, points B and C are not considered neighboring points in the map. However, these points are close to each other, as depicted in the image resulting from SLAM [29].

Frontend algorithms can be classified into two categories of feature and direct methods. Feature methods have demonstrated excellent performance compared to direct methods due to their stability under lighting variations and rotations. Several algorithms are available for feature extraction, such as SIFT, SURF, ORB [30], BRIEF, and FAST. Given the real-time requirements of VIP's navigation systems, feature extraction methods that can operate in real-time with low-cost devices are needed. ORB was chosen as the feature extraction method in the proposed method due to its superior speed compared to other methods [31][32].

There are several frameworks available for implementing V-SLAM. MonoSLAM (2007) consists of four steps: Initialization, State Prediction, Tracking, and Correction. Although MonoSLAM is a pioneering V-SLAM technique, it has a few drawbacks. For example, the complexity of the algorithm increases proportionally with the size of the environment. Another influential V-SLAM algorithm is PTAM (2007), which introduced separated and parallelized tracking and mapping threads. It was the first algorithm to separate the frontend and backend components of V-SLAM. PTAM uses FAST detectors for feature extraction, but they may struggle with extreme situations such as rotations and intensity variations. DTAM is the first algorithm to employ direct methods for map creation, providing detailed reconstruction that increases complexity. However, all the mentioned V-SLAM algorithms share a common issue: the absence of loop closing and global optimization techniques, which can lead to inaccuracies in mapping and tracking results [27]

One algorithm that supports loop closing and global optimization techniques and utilizes ORB features is ORBSLAM [32],[17]. It is the most famous and straightforward form of V-SLAM that supports real-time applications with a three-layered structure[6]. ORBSLAM is used in the proposed method since it can work with monocular cameras, run on both CPU and GPU and can work in real time [27].

They utilized ORB features for all processes to achieve real-time execution without requiring a GPU. Additionally, to support real-time processing in large environments, they employed a covisibility graph for mapping and tracking local areas independently of the global map. Another contribution of their work is real-time loop closing based on pose graph optimization and real-time camera relocation. In order to enhance efficiency, they adopted a procedure to select keyframes from the frames captured, rather than processing every frame [32].

For camera relocation and loop closing, they have used a method based on an image-to-image comparison named Bag of Words. DBoW2 [33] is one of the exemplary implementations of the Bag of Words method, which uses FAST and BRIEF feature detectors and descriptors. However, their tests showed that these detectors were incompatible with rotation and scale, limiting place recognition to fixed angle viewpoints. To address this issue, the authors of [32] [17] proposed a method based on DBoW2 [34] that uses ORB features. The entire loop closing process, including ORB feature extraction, took less than 39 milliseconds.

ORBSLAM consists of three main parts: Tracking, Local Mapping, and Loop Closure, each running on a separate thread. The tracking thread is responsible for extracting ORB features, estimating the camera pose, and inserting the new keyframes. Since a monocular camera is used in the proposed method, depth information is not available to create the map. Therefore, a process called Map Initialization is used to create an initial map for camera localization in the tracking thread.

The local mapping section processes selected keyframes to reconstruct the camera's surrounding area and performs local bundle adjustment to optimize the positions. The system searches corresponding unmatched ORB features of the new keyframe in the covisibility graph to triangulate and reconstruct the scene. Additionally, the local mapping applies a culling policy to retain high-quality points and remove redundant keyframes.

In the last part, loop closing searches for new loops when a new keyframe is received. Once a new loop is detected, both sides of the loop are connected, and duplicate points are removed. This thread performs pose graph optimization to fully optimize the map.

### Obstacle Detection:

The YOLOv4 framework [35], which is based on YOLO (You Only Look Once [36]), is used in the proposed system to detect obstacles. Yolo is a CNN network that simultaneously predicts both bounding boxes and object classes. The neural network converts the input image into an

S*S gird, where each cell containing the object's center is responsible for detecting the bounding boxes and the class of the object. The bounding boxes and their confidence scores are predicted for each cell. A confidence score specifies how much the system is confident that there is an object in a bounding box and if the bounding box is located in an actual position compared to ground truth. The width, height, center coordinates, and confidence score of the bounding box are predicted. After detecting the bounding box, classification is done to predict the object's class.

The YOLO network is made of three kinds of layers. Convolution layers, fully connected layers, and output layers. Convolution layers are responsible for extracting and processing the image features. Convolution layers are pre-trained on classification datasets such as ImageNet with lower resolutions. Fully connected layers use these features to predict bounding boxes and their positions. The output of this network is S*S*(C+B*5) tensor.

YOLOv4 is a two-stage detector with four parts in its architecture: Input, Backbone, Neck, and Head. The input part is responsible for passing the user input to the backbone which extracts feature and pass them to the neck for optimization. Finally, the head component makes the predictions. An overview of architecture is shown in figure 4.

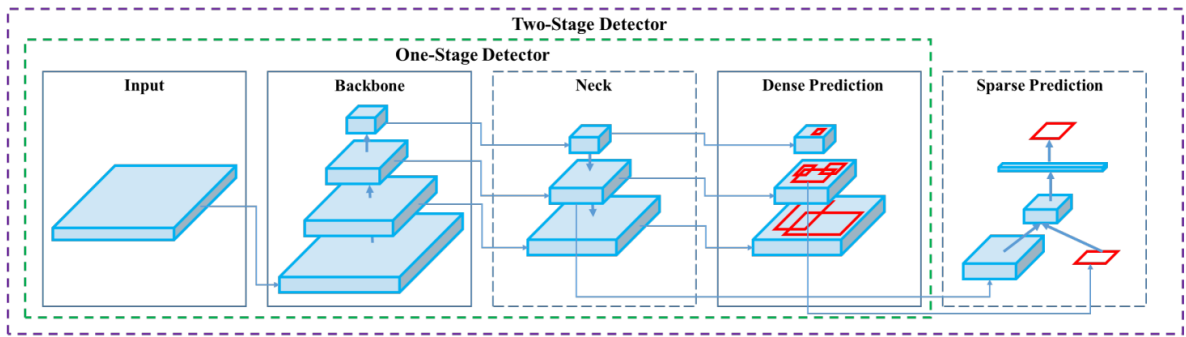

Figure 4 YOLOv4 architecture

Various tools and frameworks are available to implement each part of the mentioned structure. For example, options for the backbone include VGG16, Darknet53, or CSPDarknet53. In the neck part, FPN or PANet can be used. CSPDarknet53 is selected as the backbone due to its optimal performance. This network separates the most significant features and causes almost no reduction in the operation of the network speed. For the head part, YOLOv3 is used.

Various strategies are employed to enhance the performance and accuracy of YOLOv4, selected from the Bag of Freebies. The Bag of Freebies consists of techniques that boost accuracy without significantly increasing inference costs. These strategies often increase training costs but have no practical impact. YOLOv4 utilizes techniques such as Data Augmentation and the Semantic Distribution Bias Objective Function for BBox regression to improve its performance and accuracy.

The Bag of Specials is another set of strategies employed, particularly in the post-processing stage. While these techniques may slightly increase inference costs, they significantly improve object detection accuracy. An example is the utilization of mish activation functions instead of ReLU, which results in a notable increase in testing accuracy. Other techniques within the Bag of Specials include Cross-stage partial connections (CSP), SPP-block, and SAM-block.

### Path Following:

In the path following part, we aim to help the user stay within the SLAM-created map through three steps of obstacle distance estimation, path deviation detection, and estimating the next step in the path. To keep the concept simple, we have used a monocular camera and a SLAM-created map for all the mentioned tasks. In monocular SLAM systems, the map is created with a fixed scale during the initialization step. We load the map with a specified scale and perform our calculations based on that.

In the obstacle detection part, the position of the object's bounding box center is found. This position is then compared with the map points captured by the camera. The map point with the shortest Euclidean distance to the bounding box center is chosen and labeled as an "Object". The world coordinates of this point are obtained from the map, and its colors is changed to green, while other ones are shown in red. As the camera sees the point, the Euclidean distance between the camera and the point is continuously calculated. The system will notify the user if this distance falls below the specified threshold, which is a multiple of the scale plus the camera's current position.

To implement the path deviation part, current camera position is continuously compared to the path drawn by the SLAM. The path is created using camera centers stored in keyframes. If the Euclidean camera distance exceeds the path beyond the specified threshold, the system will notify the user to return to the path.

To estimate the next step, the 2D cross product of a vector has been used which is calculated as $(A_x * B_y) – (A_y * B_x)$. Three points of camera's position, nearest keyframe to the camera, and Kth next keyframe are used to estimate and guide the user for the next steps. To determine the direction, all coordinates are subtracted from the position of camera to establish the camera as origin. Then cross product of two point meaning keyframe and query point is calculated. If the result is negative or positive the user should go to the left or right respectively, and if it is zero, it indicates that three points are colinear, and the user should proceed straight ahead.

### Experimental Results:

In this section, the evaluation of the proposed system is presented. The proposed system run in real-time on a CPU, thanks to ORB-SLAM and YOLOv4 frameworks. Experimental results were obtained using a laptop equipped with an Intel Core i7-10750H 2.60GHz processor using Linux 20.0.4 operating system. HD User Facing Acer built-in webcam has been used to capture video test frames. The proposed system is analyzed in four major parts. First, map selection scenes are explained. After that, the results related to the performance of map creation of each scene are shown. The map's coordinates and the real world are compared in the third section. Finally, the path following part including obstacle distance estimation, path deviation, and subsequent step estimation is evaluated.

#### Experimental Map Building Scenes:

Four types of path shapes were selected for concept evaluation to build the map. This method of map scene selection is inherited from [11]. The selected path shapes include a simple straight line, an L-shape path, a U-shape path, and a square-shape path, as shown in Figure 5. These four types of paths were used to evaluate the proposed method. Additionally, the square-shape path was specifically used to assess the loop closing part of ORB-SLAM.

Map Building Performance Analysis:

To build the map using ORB features in ORB-SLAM, four parameters need to be determined: the maximum number of features, the scale factor, the number of levels in the ORB scale pyramid, and the initial FAST key-point detector threshold. These parameters are set to 2000, 1.2, 8, and 10, respectively. For each shape of the path introduced in the previous section, maps and a sample image, including the key points are shown in figure 6.

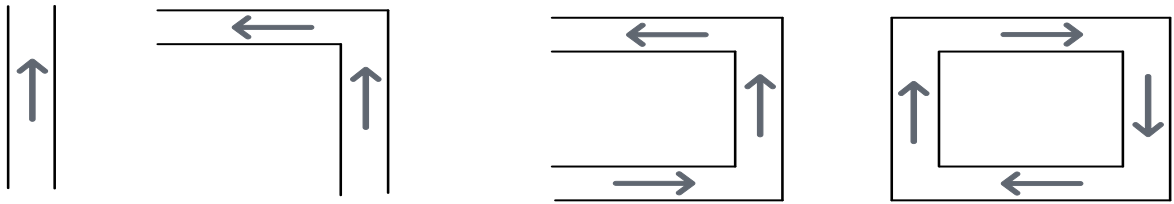

Figure 5 From left to right: A simple straight line, L-shaped, U-shaped, and square

Table 1 presents the results regarding the performance and speed of map building. This table displays the number of keyframes utilized for map construction, which is significantly lower than the total number of frames due to the real-time execution of the method. The table demonstrates that map creation, even in the worst-case scenario, takes approximately a minute, thanks to the real-time operation of ORB-SLAM.

|  | Total Frames | Key Frames | Total Map Points | Total Time(s) |
|---|---|---|---|---|
| Straight | 357 | 50 | 4770 | 21.79 |
| L-Shaped | 487 | 61 | 5498 | 35.72 |
| U-Shaped | 721 | 108 | 10498 | 46.78 |
| Square | 805 | 113 | 9743 | 58.46 |

Table 1 Map building performance results

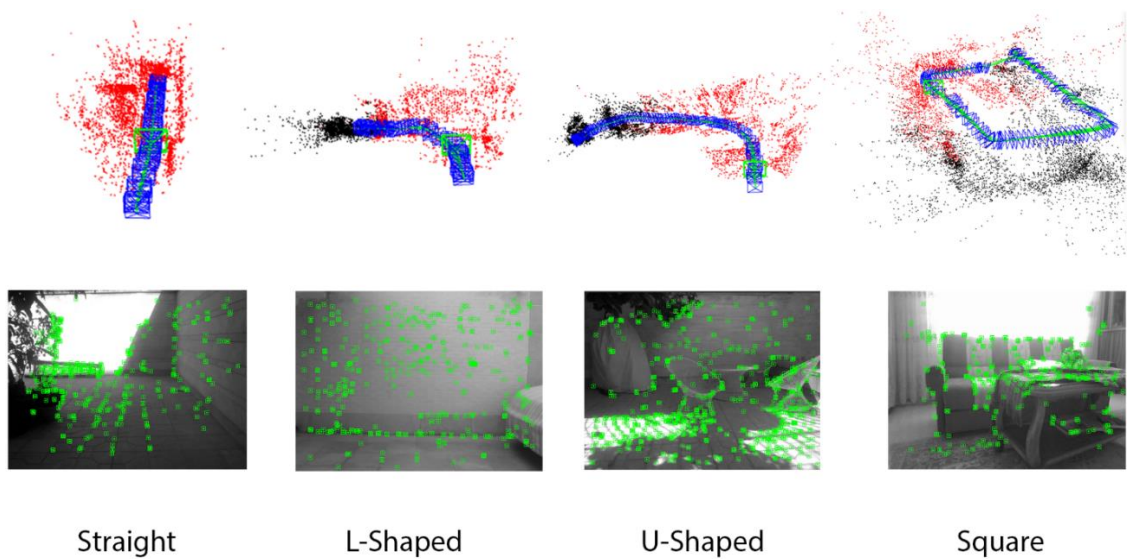

Figure 6 Experimental Maps created from selected scenes

## Map and Real-world Coordinates Mapping:

As mentioned earlier, a monocular camera is utilized to implement the V-SLAM concept. [28]Monocular SLAM systems inherently have scale ambiguity [28], meaning the exact size or measurement unit in the world map is uncertain. In these systems, the scale is defined during the map initialization and remains fixed throughout the entire process. The scale used in creating the map remains fixed throughout the testing process, and therefore, the map is loaded with the same scale during testing.

Now that the scale is consistent and unique during the map loading, map point positions are also unchanged and consistent. This attribute can be used to test the proposed system and measure the error. To calculate the error, a set of pair points called checkpoints is selected. The actual distance between these checkpoints is determined using a tape measure as a reference. The coordinates of check points are recorded, and distance between them is calculated in map units. A real distance of 0.1 units in map coordinates is measured using the tape measure as a reference unit. Subsequently, the real distances between checkpoints are estimated using the reference unit and distance of each checkpoint in map coordinates. Finally, the absolute error is calculated by comparing the actual and estimated distances. These measurements are performed for each type of map mentioned earlier. The results of these calculations are presented in table 2 to table 5. Each table contains labels assigned to each checkpoint, the distance between checkpoints in map coordinates, the estimated real-world distance, the actual distance, and the corresponding error. All real-world distances are provided in centimeters. Table 2, Table 3, Table 4, and Table 5 represent these calculations for straight, l-shaped, u-shaped, and square shape paths, respectively.

Three to four checkpoints for each map type are considered to calculate the distance in real-world and world pose. The Euclidean distance is used to calculate the world pose. Since the test cases are on a flat space, there is nearly zero variation on the Y-axis, and therefore, the Y parameter is not considered in the calculations.

| Check Points | Distance (world pose) | Approx. Distance (cm) | Actual Distance (cm) | Error (cm) |
|---|---|---|---|---|
| A | 0.36 | 248 | 240 | 8 |
| B | 0.38 | 261.8 | 252 | 9.8 |
| C | 0.21 | 144.6 | 151 | 6.4 |

*Table 2 Map and Real-world mapping, Straight Map Shape, every 68.9cm is 0.1 in the world pose*

| Check Points | Distance (world pose) | Approx. Distance (cm) | Actual Distance (cm) | Error (cm) |
|---|---|---|---|---|
| A | 0.34 | 253.6 | 261 | 7.4 |
| B | 0.26 | 193.9 | 200 | 6.1 |
| C | 0.39 | 290.9 | 280 | 10.9 |

*Table 3 Map and Real-world mapping, L Map Shape, every 74.6 cm is 0.1 in the world pose*

| Check Points | Distance (world pose) | Approx. Distance (cm) | Actual Distance (cm) | Error (cm) |
|---|---|---|---|---|
| A | 0.21 | 139.8 | 150 | 10.2 |
| B | 0.31 | 206.4 | 210 | 3.6 |
| C | 0.30 | 199.8 | 210 | 10.2 |

*Table 4 Map and Real-world mapping, U Map Shape, every 66.6 cm is 0.1 in the world pose*

| Check Points | Distance (world pose) | Approx. Distance (cm) | Actual Distance (cm) | Error (cm) |
|---|---|---|---|---|
| A | 0.22 | 141.9 | 130 | 11.9 |
| B | 0.29 | 187 | 195 | 8 |
| C | 0.30 | 193.5 | 200 | 6.5 |

*Table 5 Map and Real-world mapping, Square Map Shape, every 64.5 cm is 0.1 in the world pose*

Based on the results reported in Tables 2 to 5, the mean absolute error is 8.25 cm.

### Path Following Experiments Analysis:

For obstacle distance estimation and path deviation, a threshold of 60 cm is chosen. This threshold is converted to the world pose using the equations found for every map. If the user's current position deviates from the predefined map and exceeds the threshold, the system notifies the user. In the step prediction phase, each keyframe is compared to the next five keyframes. The accuracy of this process is calculated as the ratio of True Positives (TP) to the total number of keyframes. The accuracy for each map type is calculated and presented in Table 6.

| Map Type | Accuracy |
|---|---|
| Straight | 100 % |
| L-Shaped | 80.35 % |
| U-Shaped | 83.80 % |
| Square | 87.03 % |

*Table 6 Accuracy of Next Step Prediction*

Based on the above table, it can be concluded that the overall accuracy of our system in the next step prediction section is 87.795 percent.

## Conclusions:

In this paper, we propose a navigation system based on ORB-SLAM and YOLO algorithm version 4, utilizing a monocular camera to assist visually impaired individuals in navigating a predefined path. The proposed method comprises three steps: obstacle distance estimation, path deviation detection, and next-step prediction. It is designed to function both outdoors and indoors, without relying on GPS. The method achieves a mapping accuracy of 8.5 cm between world pose coordinates and the real world. Moreover, it demonstrates an 87 percent accuracy in next-step predictions and operates at a speed of 30 frames per second.

## References:

[1] REAL, S. & ARAUJO, A. 2019. Navigation systems for the blind and visually impaired: Past work, challenges, and open problems. *Sensors,* 19**,** 3404.
[2] KURIAKOSE, B., SHRESTHA, R. & SANDNES, F. E. 2022. Tools and technologies for blind and visually impaired navigation support: a review. *IETE Technical Review,* 39**,** 3-18.
[3] LV, Z., LI, J., LI, H., XU, Z. & WANG, Y. 2021. Blind travel prediction based on obstacle avoidance in indoor scene. *Wireless Communications and Mobile Computing,* 2021.
[4] BAI, J., LIAN, S., LIU, Z., WANG, K. & LIU, D. 2018. Virtual-blind-road following-based wearable navigation device for blind people. *IEEE Transactions on Consumer Electronics,* 64**,** 136-143.
[5] KHAN, S., NAZIR, S. & KHAN, H. U. 2021. Analysis of navigation assistants for blind and visually impaired people: A systematic review. *IEEE Access,* 9**,** 26712-26734.

[6] CHEN, Z., LIU, X., KOJIMA, M., HUANG, Q. & ARAI, T. 2021. A wearable navigation device for visually impaired people based on the real-time semantic visual slam system. *Sensors,* 21**,** 1536.
[7] SÁNCHEZ, J., ESPINOZA, M., DE BORBA CAMPOS, M. & MERABET, L. B. Enhancing orientation and mobility skills in learners who are blind through video gaming.  Proceedings of the 9th ACM Conference on Creativity & Cognition, 2013. 353-356.
[8] CHAUDARY, B., POHJOLAINEN, S., AZIZ, S., ARHIPPAINEN, L. & PULLI, P. 2021. Teleguidance-based remote navigation assistance for visually impaired and blind people—Usability and user experience. *Virtual Reality***,** 1-18.
[9] BHOWMICK, A. & HAZARIKA, S. M. 2017. An insight into assistive technology for the visually impaired and blind people: state-of-the-art and future trends. *Journal on Multimodal User Interfaces,* 11**,** 149-172.
[10] KHENKAR, S., ALSULAIMAN, H., ISMAIL, S., FAIRAQ, A., JARRAYA, S. K. & BEN-ABDALLAH, H. 2016. ENVISION: assisted navigation of visually impaired smartphone users. *Procedia Computer Science,* 100**,** 128-135.
[11] XIE, Z., LI, Z., ZHANG, Y., ZHANG, J., LIU, F. & CHEN, W. 2022. A Multi-Sensory Guidance System for the Visually Impaired Using YOLO and ORB-SLAM. *Information,* 13**,** 343.
[12] MERUGU, S. & GHINEA, G. 2022. A Review of Some Assistive Tools and their Limitations for Visually Impaired. *Helix-The Scientific Explorer| Peer Reviewed Bimonthly International Journal,* 12**,** 1-9.
[13] MORE, P. R., RAUT, P. S. & WAGHMODE, P. M. 2021. Virtual eye for visually blind people. *INTERNATIONAL JOURNAL,* 5.
[14] DÍAZ-TORO, A. A., CAMPAÑA-BASTIDAS, S. E. & CAICEDO-BRAVO, E. F. 2021. Vision-based system for assisting blind people to wander unknown environments in a safe way. *Journal of Sensors,* 2021.
[15] DERPANIS, K. G. 2010. Overview of the RANSAC Algorithm. *Image Rochester NY,* 4**,** 2-3.
[16] Bai, J., Lian, S., Liu, Z., Wang, K. and Liu, D., 2017. Smart guiding glasses for visually impaired people in indoor environment. *IEEE Transactions on Consumer Electronics*, *63*(3), pp.258-266.
[17] MUR-ARTAL, R. & TARDÓS, J. D. 2017. Orb-slam2: An open-source slam system for monocular, stereo, and rgb-d cameras. *IEEE transactions on robotics,* 33**,** 1255-1262.
[18] HAKIM, H. & FADHIL, A. Indoor Wearable Navigation System Using 2D SLAM Based on RGB-D Camera for Visually Impaired People.  Proceedings of First International Conference on Mathematical Modeling and Computational Science, 2021. Springer, 661-672.
[19] HAKIM, H. & FADHIL, A. Navigation system for visually impaired people based on RGB-D camera and ultrasonic sensor.  Proceedings of the International Conference on Information and Communication Technology, 2019. 172-177.
[20] KOLHATKAR, C. & WAGLE, K. 2021. Review of SLAM algorithms for indoor mobile robot with LIDAR and RGB-D camera technology. *Innovations in electrical and electronic engineering***,** 397-409.
[21] OTSU, N. 1979. A threshold selection method from gray-level histograms. *IEEE transactions on systems, man, and cybernetics,* 9**,** 62-66.
[22] PASZKE, A., CHAURASIA, A., KIM, S. & CULURCIELLO, E. 2016. Enet: A deep neural network architecture for real-time semantic segmentation. *arXiv preprint arXiv:1606.02147*.
[23] DUH, P.-J., SUNG, Y.-C., CHIANG, L.-Y. F., CHANG, Y.-J. & CHEN, K.-W. 2020. V-eye: A vision-based navigation system for the visually impaired. *IEEE Transactions on Multimedia,* 23**,** 1567-1580.
[24] SATTLER, T., LEIBE, B. & KOBBELT, L. Fast image-based localization using direct 2d-to-3d matching.  2011 International Conference on Computer Vision, 2011. IEEE, 667-674.

[25] POHLEN, T., HERMANS, A., MATHIAS, M. & LEIBE, B. Full-resolution residual networks for semantic segmentation in street scenes.  Proceedings of the IEEE conference on computer vision and pattern recognition, 2017. 4151-4160.

[26] Alsadik, B. and Karam, S., 2021. The simultaneous localization and mapping (SLAM)-An overview. *Surv. Geospat. Eng. J*, *2*, pp.34-45.

[27] Macario Barros, A., Michel, M., Moline, Y., Corre, G. and Carrel, F., 2022. A comprehensive survey of visual slam algorithms. *Robotics*, *11*(1), p.24.

[28] Gao, X. and Zhang, T., 2021. *Introduction to Visual SLAM: From Theory to Practice*. Springer Nature.

[29] Cadena, C., Carlone, L., Carrillo, H., Latif, Y., Scaramuzza, D., Neira, J., Reid, I. and Leonard, J.J., 2016. Past, present, and future of simultaneous localization and mapping: Toward the robust-perception age. *IEEE Transactions on robotics*, *32*(6), pp.1309-1332.

[30] Rublee, E., Rabaud, V., Konolige, K. and Bradski, G., 2011, November. ORB: An efficient alternative to SIFT or SURF. In *2011 International conference on computer vision* (pp. 2564-2571). Ieee.

[31] Karami, E., Prasad, S. and Shehata, M., 2017. Image matching using SIFT, SURF, BRIEF and ORB: performance comparison for distorted images. *arXiv preprint arXiv:1710.02726*.

[32] Mur-Artal, R., Montiel, J.M.M. and Tardos, J.D., 2015. ORB-SLAM: a versatile and accurate monocular SLAM system. *IEEE transactions on robotics*, *31*(5), pp.1147-1163.

[33] Gálvez-López, D. and Tardos, J.D., 2012. Bags of binary words for fast place recognition in image sequences. *IEEE Transactions on Robotics*, *28*(5), pp.1188-1197.

[34] Mur-Artal, R. and Tardós, J.D., 2014, May. Fast relocalisation and loop closing in keyframe-based SLAM. In *2014 IEEE International Conference on Robotics and Automation (ICRA)* (pp. 846-853). IEEE.

[35] Bochkovskiy, A., Wang, C.Y. and Liao, H.Y.M., 2020. Yolov4: Optimal speed and accuracy of object detection. *arXiv preprint arXiv:2004.10934*.

[36] Redmon, J., Divvala, S., Girshick, R. and Farhadi, A., 2016. You only look once: Unified, real-time object detection. In *Proceedings of the IEEE conference on computer vision and pattern recognition* (pp. 779-788).